\PassOptionsToPackage{table,xcdraw,usenames,dvipsnames}{xcolor}
\documentclass[11pt]{article}

\usepackage[preprint]{acl}

\usepackage{times}
\usepackage{latexsym}

\usepackage[T1]{fontenc}
\usepackage[utf8]{inputenc}

\usepackage{microtype}

\usepackage{inconsolata}

\usepackage{graphicx}

\usepackage{hyperref}
\usepackage{url}
\usepackage{amsthm}
\usepackage{hyperref}

\usepackage[most]{tcolorbox}
\usepackage{wrapfig}
\usepackage{multirow}
\usepackage{graphicx,xcolor,float}
\usepackage{subfigure}
\usepackage{threeparttable}
\usepackage[ruled,vlined]{algorithm2e}
\usepackage{colortbl}
\usepackage{color}
\usepackage{multirow}
\usepackage{tabularx}
\usepackage{float}
\usepackage{graphicx}
\usepackage{booktabs}
\usepackage{arydshln}
\usepackage{enumitem}
\usepackage{wrapfig}
\usepackage{caption}
\usepackage{graphicx}
\usepackage{makecell}
\usepackage{float} 
\usepackage{mathtools}
\usepackage{subfigure}
\usepackage{bbding}
\usepackage{makecell}
\usepackage{tabularx}
\usepackage{amssymb,mathrsfs,amsmath}
\usepackage{pifont}
\usepackage{bbm}

\usepackage{listings}

\newcommand{\ourmethod}{{\fontfamily{lmtt}\selectfont \textbf{EvoRoute}}\xspace}
\newcommand{\llmname}[1]{{\texttt{#1}}}

\definecolor{bittersweet}{rgb}{1.0, 0.44, 0.37}
\definecolor{mygreen}{rgb}{0.29, 0.7, 0.48}
\usepackage{pifont}% http://ctan.org/pkg/pifont for xmark and cmark
\definecolor{demphcolor}{RGB}{144,144,144}

\definecolor{mygray}{gray}{0.4}
\definecolor{autopurple}{HTML}{7030A0}
\definecolor{dyna_yellow}{HTML}{BF9000}
\definecolor{adaptive_blue}{HTML}{0070C0}
\definecolor{darksalmon}{rgb}{0.91, 0.59, 0.48}
\definecolor{emerald}{rgb}{0.31, 0.78, 0.47}
\definecolor{green(pigment)}{rgb}{0.0, 0.65, 0.31}
\definecolor{amaranth}{rgb}{0.9, 0.17, 0.31}
\definecolor{iris}{rgb}{0.35, 0.31, 0.81}
\definecolor{uu}{rgb}{0.95, 0.51, 0.51}
\definecolor{spirodiscoball}{rgb}{0.06, 0.75, 0.99}
\usepackage{svg}

\usepackage{cleveref}
\SetKwInOut{Input}{Input}\SetKwInOut{Output}{Output}

\title{EvoRoute: Experience-Driven Self-Routing LLM Agent Systems}

\author{Guibin Zhang, Haiyang Yu, Kaiming Yang, Bingli Wu, Fei Huang, Yongbin Li, Shuicheng YAN}

\author{
 \textbf{Guibin Zhang\textsuperscript{1}},
 \textbf{Haiyang Yu\textsuperscript{2}},
 \textbf{Kaiming Yang\textsuperscript{1}},
 \textbf{Bingli Wu\textsuperscript{2}},
\\
 \textbf{Fei Huang\textsuperscript{2}},
 \textbf{Yongbin Li\textsuperscript{2$\dagger$}},
 \textbf{Shuicheng Yan\textsuperscript{1$\dagger$}}
\\
{\small
 \textsuperscript{1}National University of Singapore,
 \textsuperscript{2}Tongyi Lab,
 \textsuperscript{$\dagger$}Corresponding Authors
 }
\\
 \small{
   \textbf{Main Contact:} \href{mailto:guibinz@outlook.com}{\texttt{guibinz@outlook.com}}
 }
}

\begin{document}
\maketitle
\begin{abstract}
Complex agentic AI systems, powered by a coordinated ensemble of Large Language Models (LLMs), tool and memory modules, have demonstrated remarkable capabilities on intricate, multi-turn tasks. However, this success is shadowed by prohibitive economic costs and severe latency, exposing a critical, yet underexplored, trade-off. We formalize this challenge as the \textbf{Agent System Trilemma}: the inherent tension among achieving state-of-the-art performance, minimizing monetary cost, and ensuring rapid task completion.
To dismantle this trilemma, we introduce \ourmethod, a self-evolving model routing paradigm that transcends static, pre-defined model assignments.  Leveraging an ever-expanding knowledge base of prior experience, \ourmethod dynamically selects Pareto-optimal LLM backbones at each step, balancing accuracy, efficiency, and resource use, while continually refining its own selection policy through environment feedback. Experiments on challenging agentic benchmarks such as GAIA and BrowseComp+ demonstrate that \ourmethod, when integrated into off-the-shelf agentic systems, not only sustains or enhances system performance but also reduces execution cost by up to $80\%$ and latency by over $70\%$. %Our codes are available at \url{https://anonymous.4open.science/r/EvoRoute}.
\end{abstract}

\section{Introduction}
% 随着llm强大 agentic ai system feature on：多个异构模型调用 tool memory
% 在各类复杂任务上更加成功

As {Large Language Model (LLM)-powered agents} continue to demonstrate increasingly advanced cognitive capabilities, spanning perception~\citep{driess2023palm-e,zheng2023steve,wei2024editable}, planning~\citep{zhu2024knowagent,erdogan2025plan-and-act}, reasoning~\citep{putta2024agentq,masterman2024landscape}, and action~\citep{li2024embodied,yang2024embodied}, LLM-driven agentic AI systems have achieved remarkable performance across a range of highly complex, multi-turn tasks, including machine learning engineering~\citep{chan2025mlebenchevaluatingmachinelearning}, multi-hop information searching~\citep{mialon2023gaia}, report generation~\citep{chen2025xbenchtrackingagentsproductivity}, and GitHub issue repair~\citep{jimenez2024swebenchlanguagemodelsresolve}.

\begin{figure}
\vspace{-0.38cm}
  \begin{center}
    \includegraphics[width=\columnwidth]{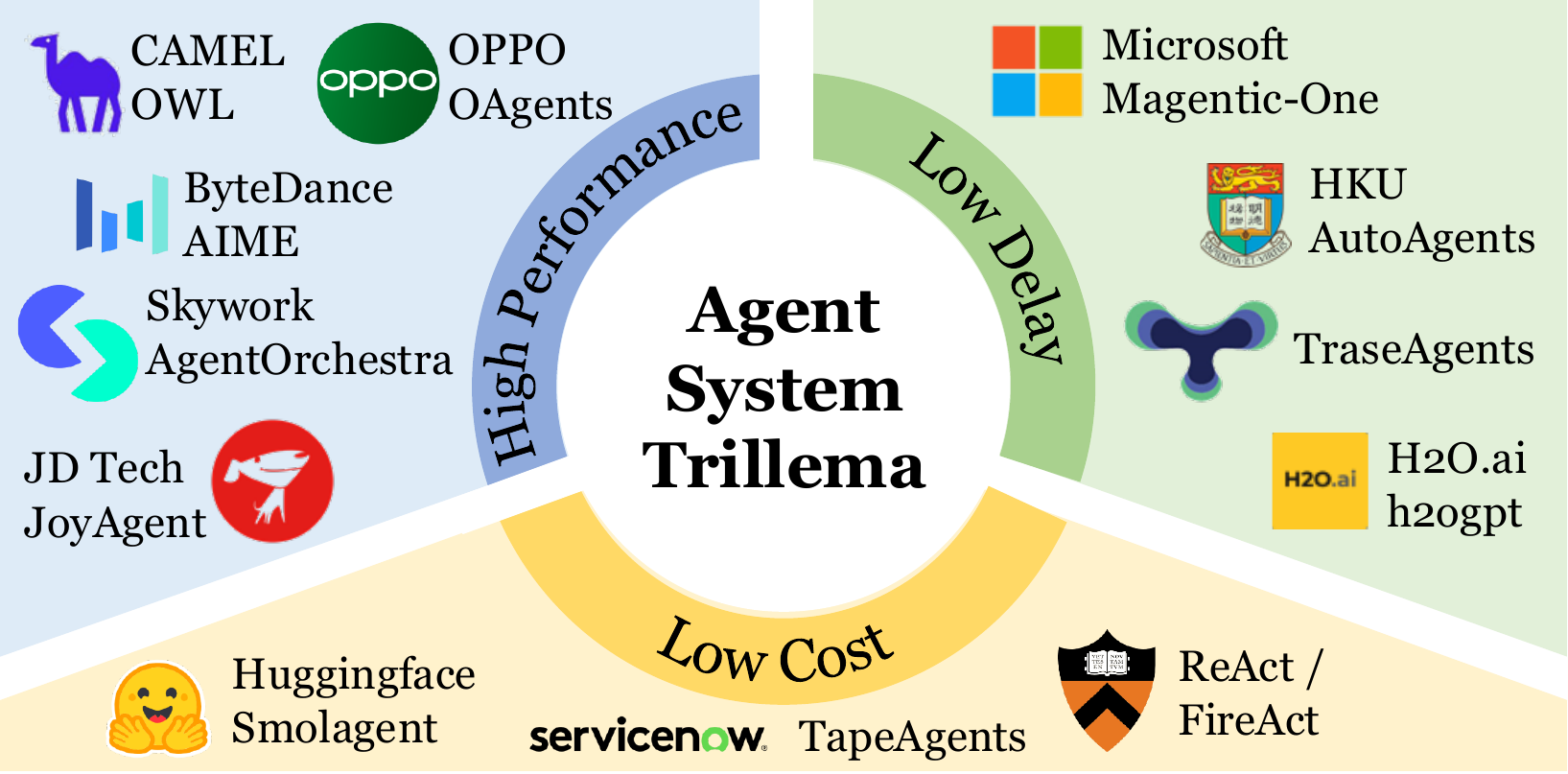}
  \end{center}
  \vspace{-0.2cm}
  \caption{The agent system trilemma. Existing (deep research) agentic systems excel in certain aspects, yet none of which can fulfill the three characteristics spontaneously.}
  \label{fig:intro}
  \vspace{-0.4cm}
\end{figure}

Moving beyond single-agent frameworks, these agentic systems are distinguished by the coordinated operation of multiple heterogeneous models (potentially developed by different institutions or specialized for distinct domains) alongside sophisticated tool invocation workflows and memory modules~\citep{chen2025optimizingmodelselectioncompound,chaudhry2025resourceefficientcompoundaisystems}. Prominent recent instances include AgentOrchestra~\citep{zhang2025agentorchestrahierarchicalmultiagentframework} from Skywork AI, which achieved a state-of-the-art (SOTA) open-source score of $70\%+$ on GAIA~\citep{mialon2023gaia}, and ML-master~\citep{liu2025mlmasteraiforaiintegrationexploration}, which obtained $29.3\%$ on MLE-Bench~\citep{chan2025mlebenchevaluatingmachinelearning}, both of which involve the utilization of cutting-edge models (\textit{e.g.}, OpenAI GPT-4.1~\citep{gpt41}, Gemini-2.5-Pro~\citep{comanici2025gemini}).

Despite their impressive performance, several underexplored evaluation dimensions raise concerns about the practical deployment of such systems. Empirical studies have revealed that certain general-purpose agentic systems, such as OWL-Workforce~\citep{hu2025owloptimizedworkforcelearning}, which achieved open-source SOTA on GAIA in June 2025, incur substantial execution costs, averaging up to \$3 per task. In some cases, even relatively simple queries may entail up to 40 minutes of execution, despite ultimately arriving at the correct answer. Similarly, R\&D Agent~\citep{yang2025rdagent} failed to complete the 75 data science tasks specified by MLE-Bench within a 24-hour window. In summary, the \ding{168} \textbf{efficiency} (\textit{i.e.}, task completion speed) and the \ding{171} \textbf{cost} (\textit{i.e.}, cumulative expenditure on LLM invocations and tool usage) of compound agentic systems often fall short of their promising \ding{169} \textbf{performance} metrics, as shown in \Cref{fig:intro}.

To explicitly formalize this challenge, we draw inspiration from the classical ``impossible trinity'' theory in economics~\citep{aizenman2013impossible} and blockchain~\citep{koutsoupakis2021blockchain-trinity}, where it is deemed infeasible for a system to simultaneously achieve all three desirable properties (\textit{e.g.}, scalability, security, and decentralization in the blockchain context). Analogously, we propose the \textbf{Agent System Trilemma} to characterize the inherent trade-offs in complex agentic systems. Specifically, when functioning as general-purpose AI assistants, such systems often face an intrinsic tension among three core objectives: \textbf{performance} (\textit{i.e.}, task success and accuracy), \textbf{efficiency} (\textit{i.e.}, time or steps required to complete tasks), and \textbf{cost} (\textit{i.e.}, computational and monetary resources consumed).

Is it truly impossible to devise an approach capable of overcoming this trilemma and achieving a \textit{tri-optimal} agentic system? We posit that the answer is affirmative. A particularly promising direction lies in {model routing}~\citep{hu2024routerbenchbenchmarkmultillmrouting}, \textit{i.e.}, intelligently selecting the most suitable LLM agent from a pool of candidate models given a task query.
However, existing model routing paradigms remain insufficient to break the trilemma.  The first, \ding{228} \textbf{single model routing}, \textit{i.e.}, assigning one model to handle the entire task~\citep{feng2024graphroutergraphbasedrouterllm,chen2024routerdc}, works for atomic tasks but struggles with complex agentic workflows, where sub-tasks like web browsing, coding, and summarization often require distinct model capabilities~\citep{chen2025optimizingmodelselectioncompound,frick2025prompttoleaderboard}.
The second paradigm, \ding{228} \textbf{multi-agent routing}, formally defined in~\cite{yue2025masrouter,chen2025optimizingmodelselectioncompound}, selects different models for different agent roles within a multi-agent system. Although this approach aligns more closely with the structure of agentic systems, it has so far been limited to relatively simple, single-step reasoning scenarios (\textit{e.g.}, math reasoning and code generation), and relies heavily on large-scale trajectories to model agent behavior, making it ill-suited for dynamic, multi-turn agentic scenarios.

To confront this trilemma, we introduce \ourmethod, a tri-optimal, self-evolving model routing paradigm for general-purpose agentic systems. Specifically, rather than committing to a single model or a fixed multi-agent configuration, \ourmethod operates at the granularity of individual sub-tasks within a complex workflow. Before executing each step, it dynamically selects the most judicious LLM by: \textbf{\ding{182} retrieval}, performing a multi-faceted retrieval to identify historically analogous sub-task executions from an evolving knowledge base; \textbf{\ding{183} filtration}, distilling a Pareto-optimal set of candidate models, \textit{i.e.}, those that are not dominated across the axes of cost, efficiency, and performance; and \textbf{\ding{184} selection}, leveraging a lightweight decision model to make the final selection based on this rich, context-aware statistical evidence. 

The ``self-evolving'' nature of \ourmethod is realized through a dual-phase operational design: during the Optimization Phase, the system engages in a tree-based exploration, sampling multiple trajectories for a given task to proactively populate and diversify its knowledge base on model behaviors; conversely, during the Inference Phase, it leverages this accumulated wisdom to pursue a single, optimized execution path, ensuring rapid and cost-effective task completion. This adaptive, experience-driven approach allows \ourmethod to continuously refine its routing strategy, evolving toward breaking the constraints of the Agent System Trilemma.

Our contributions can be summarized as follows:

\vspace{-0.5em}
\begin{itemize}[leftmargin=2em,itemsep=-0.1em]
\item[\ding{182}] \textbf{Problem Formulation.} We formalize the critical bottleneck in current agentic systems as the \textit{Agent System Trilemma}, an intrinsic trade-off among performance, cost, and efficiency, and introduce a viable approach to mitigate the issue empirically.
\item[\ding{183}] \textbf{Technical Solution.} We propose \ourmethod, a novel self-evolving routing paradigm that dismantles this trilemma through fine-grained model selection, which combines multi-faceted retrieval with Pareto-optimality selection to make resource-aware model routing.
\item[\ding{184}] \textbf{Empirical Validation.} We conduct extensive experiments on five challenging benchmarks, including GAIA and BrowseComp+, and demonstrate that \ourmethod can outperform vanilla agent systems by up to $10.3\%$ while incurring only $\sim20\%$ of the cost and achieving nearly $3\times$ faster execution.
\end{itemize}

\section{Related Work}
\vspace{-0.3em}
\paragraph{Agentic AI Systems} Contemporary multi-agent systems can be broadly categorized by their level of automation into three classes:
\ding{110} \textbf{Handcrafted}, where the entire system configuration (\textit{e.g.}, LLM backbone, prompting strategies, and communication protocols) is manually specified represented by AutoGen~\citep{autogen}, AutoGPT~\citep{autogpt}, Camel~\citep{NeurIPS2023_Agent-SoM}, and ChatDev~\citep{software-dev};
\ding{110} \textbf{Partially Automated}, which automate specific system components: for example, AutoAgent~\citep{arXiv2023_AutoAgents}, LLMSelector~\citep{chen2025optimizingmodelselectioncompound}, and MasRouter~\citep{yue2025masrouter} automate agent role assignment; DsPy~\citep{khattab2023dspy} and TextGrad~\citep{yuksekgonul2024textgradautomaticdifferentiationtext} optimize prompt design; GPTSwarm~\citep{zhuge2024gptswarm} and G-Designer~\citep{zhang2024g-designer} adaptively construct inter-agent topologies;
\ding{110} \textbf{Fully Automated}, where all modules within the system are autonomously designed and evolved~\citep{hu2024adas,zhang_aflow_2024,zhang2025maas,wu2025optimas,nie2025weakforstrongtrainingweakmetaagent,gao2025flowreasonerreinforcingquerylevelmetaagents,zhang2025evoflow}. 

\vspace{-0.3em}
\paragraph{LLM \& Agent Routing} 
Leveraging multiple models via intelligent routing has emerged as a central paradigm in modern AI and ML, aiming to exploit complementary model capabilities to enhance task performance while potentially optimizing computational costs~\citep{srivatsa2024harnessingpowermultipleminds}. Router-based approaches, which constitute the primary focus of this work, learn to assign each query to the most appropriate model~\citep{hu2024routerbenchbenchmarkmultillmrouting}. 
Classical approaches can be categorized as: (I) neural network-based routers trained on performance or cost signals, including {LLM-Blender}, which aggregates outputs from top-$k$ models selected via pairwise comparisons~\citep{blender},  {ZOOTER}, which enhances router training with reward-guided, tag-based label augmentation~\citep{lu2023routingexpertefficientrewardguided}, and others~\citep{ong2024routellmlearningroutellms,feng2024graphroutergraphbasedrouterllm,zhang2025capabilityinstructiontuningnew}; (II) cluster-based methods, including {UniRoute}~\citep{jitkrittum2025universalmodelroutingefficient}, {BEST-Route}~\citep{ding2025bestroute}, Avengers~\citep{zhang2025avengerssimplerecipeuniting} and also the baselines introduced in {RouterBench}~\citep{hu2024routerbenchbenchmarkmultillmrouting}. 
Despite their success, these methods predominantly focus on single-turn responses and are evaluated on relatively simple benchmarks (\textit{e.g.}, {Chatbot Arena}~\citep{chiang2024chatbotarenaopenplatform}, {MATH}~\citep{hendrycksmath2021}), and they have yet to achieve fine-grained routing in more complex scenarios such as deep research tasks.

\section{Preliminary}
\vspace{-0.5em}
In this section, we provide a general definition of LLM-based agentic AI systems and their operational workflow, and then formally define the objective of the model routing within this context.

\vspace{-0.5em}
\paragraph{Notations.} We consider a complex agentic AI system, \( \mathcal{M} \), designed to resolve a user-issued query \( \mathcal{Q} \) through a multi-step workflow, which is typically composed of a set of specialized agents or roles, \( \mathcal{I} = \{1, 2, \ldots, N\} \), which operate sequentially under a turn-based protocol where a scheduler determines the active agent at each time step. The workflow can involve a series of actions ranging from task decomposition to the execution of specific sub-tasks, formally defined as:
\begin{equation}
\mathcal{M} = \bigl\langle \mathcal{I}, \mathcal{L}, \phi, \mathcal{S}, \mathcal{T}, \mathcal{A}, \Psi, \mu, \mathcal{Q} \bigl\rangle,
\end{equation}
where $\mathcal{I} = \{1, 2, \ldots, N\}$ denotes the set of agent roles (\textit{e.g.}, web-browser, coder), and $\mathcal{L}$ represents the pool of available LLM backbones. Each agent $i \in \mathcal{I}$ is statically assigned an LLM via a mapping $\phi: \mathcal{I} \to \mathcal{L}$. The system state is maintained in $\mathcal{S}$, typically implemented as a shared memory or scratchpad. The agentic workflow may invoke a set of external tools $\mathcal{T}$, such as code interpreters~\citep{self-collab-codegen} or web search APIs~\citep{jimenez2024swebenchlanguagemodelsresolve}. The full action space $\mathcal{A}$ includes both natural language actions and tool invocations, formally $\mathcal{A} = \mathcal{A}_{\text{lang}} \cup \{ \texttt{use\_tool}(T, \text{args}) \mid T \in \mathcal{T} \}$. The transition dynamics of the system are governed by $\Psi(s_{t+1} \mid s_t, a_t)$, while the scheduler $\mu(t) \in \mathcal{I}$ selects the active agent at each time step $t$.

\begin{figure*}[!t]
\centering
\includegraphics[width=\linewidth]{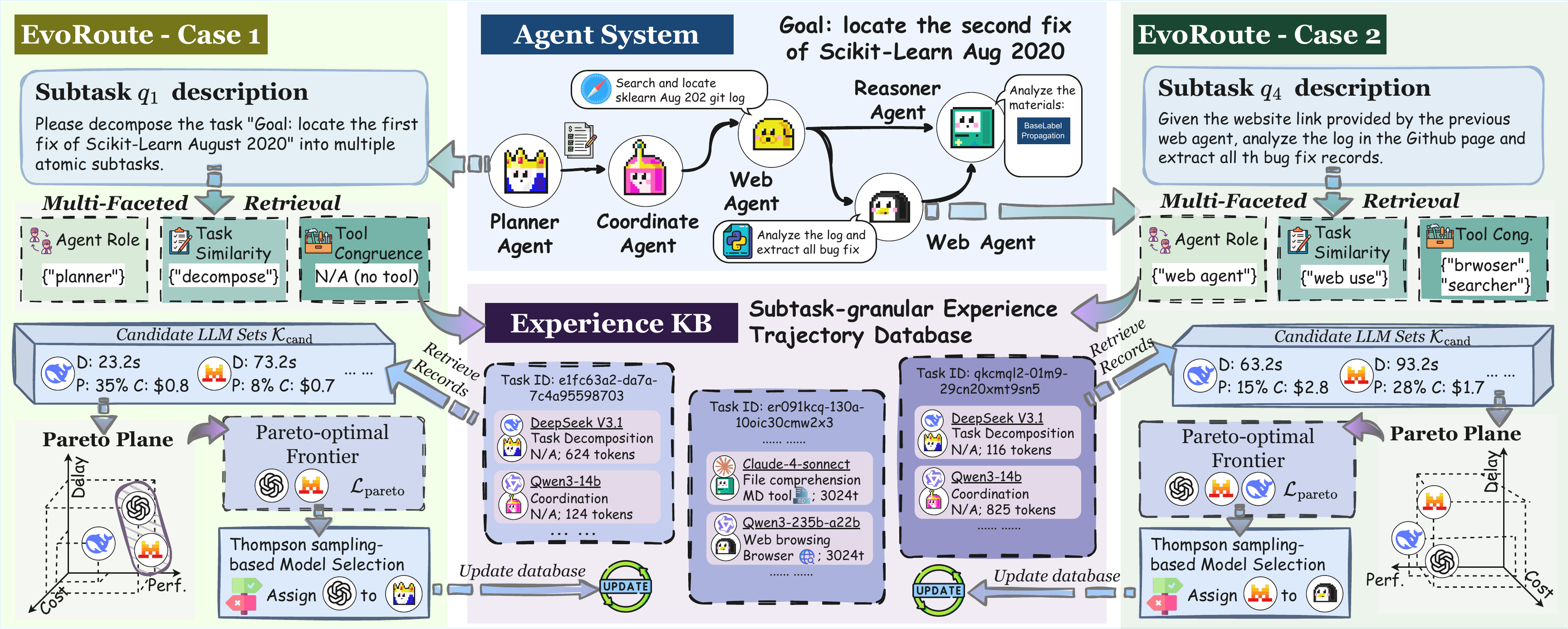}
\vspace{-1.8em}
\caption{The overview of our proposed \ourmethod.}
\label{fig:framework}
\vspace{-1.3em}
\end{figure*}

At each step \( t \), the active agent \( i = \mu(t) \) generates an action \( a_t \). This action is produced by its policy \( \pi_i \), which is instantiated by its designated LLM, \( \phi(i) \in \mathcal{L} \). The policy takes into account the current state \( s_t \), the overall query \( \mathcal{Q} \), and the relevant interaction history \( \mathcal{H}_t \):
\begin{equation}
\begin{gathered}
a_t \sim \pi_i(s_t, \mathcal{H}_t, \mathcal{Q}),\; i = \mu(t),\;\pi_i \leftarrow \phi(i).
\end{gathered}
\end{equation}
The full execution trajectory of the system is thus a sequence of states and actions:
\begin{equation}
\tau = (s_0, a_0, s_1, a_1, \ldots, s_T),
\end{equation}
where \( T \) is the terminal step. The final answer to query \( \mathcal{Q} \) is synthesized from the information aggregated throughout this trajectory \( \tau \).

A critical and often implicit characteristic of these established agentic systems is that the agent-to-model mapping, \( \phi \), is \textit{static and human-predefined}. For instance, a \texttt{Web-Browser} agent is hard-coded to use \llmname{GPT-4o} in OWL-Workforce, and \llmname{o3-mini} for a \texttt{planner} agent. Other mainstream frameworks also follow similar patterns, including Microsoft's Magentic-one~\citep{fourney2024magenticonegeneralistmultiagentsolving}, Skywork AI's AgentOrchestra~\citep{zhang2025agentorchestrahierarchicalmultiagentframework}, Bytedance's AIME~\citep{shi2025aimefullyautonomousmultiagentframework}, and Tencent's CK-Pro~\citep{fang2025cognitivekernelproframeworkdeep}. This rigid design choice, while straightforward, lacks adaptability and fails to account for the varying difficulty and nature of sub-tasks.

\vspace{-0.5em}
\paragraph{Objective Formulation.}
This paper is dedicated to transcending the limitations of a static agent-to-model mapping by learning a dynamic routing policy, \( \rho^* \), that optimally navigates the Agent System Trilemma. We formulate this as the following multi-objective optimization problem:
\begin{equation}
\small
\label{eq:objective}
\begin{gathered}
\rho^* = \underset{\rho}{\text{argmax}} ( \mathbb{E}_{\tau \sim \rho}[\mathbb{P}(\tau)], -\mathbb{E}_{\tau \sim \rho}[\mathbb{C}(\tau)], -\mathbb{E}_{\tau \sim \rho}[\mathbb{D}(\tau)] ),
\end{gathered}
\end{equation}
where \( \rho \) is the dynamic routing policy that selects an LLM \( l_t \in \mathcal{L} \) for the active agent at each step \( t \). The expectation \( \mathbb{E}_{\tau \sim \rho} \) is taken over the distribution of execution trajectories \( \tau \) generated under this policy. For each trajectory, we measure three key metrics: \ding{169} \textbf{performance} \( \mathbb{P}(\tau) \), the task success score; \ding{171} \textbf{cost} \( \mathbb{C}(\tau) \), the cumulative monetary and computational expenditure; and \ding{168} \textbf{efficiency} \( \mathbb{D}(\tau) \), the total wall-clock execution time. The pursuit of \( \rho^* \) is therefore equivalent to finding a Pareto-optimal solution that maximizes performance while concurrently minimizing cost and delay.

\vspace{-0.3em}
\section{Methodology}
\vspace{-0.5em}
\subsection{Self-Evolving Experience Base}
\vspace{-0.4em}
As shown in \Cref{fig:framework}, the cornerstone of \ourmethod's adaptive intelligence lies in its ability to accumulate and leverage past experience. This is enabled by a self-evolving experience base, denoted as \( \mathcal{K} \), which meticulously logs execution data at the step level. We leverage an exploration strategy to address the cold start issue of $\mathcal{K}$, with details in \Cref{app:cold}.  Upon the completion of an agentic workflow for a query \( \mathcal{Q} \), the resulting trajectory \( \tau = (s_0, a_0, \ldots, s_T) \) is retrospectively dissected. For each step \( t \in \{0, \ldots, T-1\} \), we extract and persist a detailed record \( \mathcal{R}_t \) that encapsulates the full context and outcome of that specific action.
Formally, each record \( \mathcal{R}_t \) is a structured tuple containing multi-faceted information essential for navigating the trilemma:
\begin{equation}
\label{eq:record_tuple}
\mathcal{R}_t = \langle i_t, l_t, q_t, \mathbf{e}_t, T_t, c_t, d_t, \sigma_t, \mathbb{P}(\tau) \rangle,
\end{equation}
where \( i_t = \mu(t) \in \mathcal{I} \) denotes the identifier of the active agent role (e.g., coder or coordinator), while \( l_t \in \mathcal{L} \) specifies the particular LLM backbone selected for that role at step \( t \). The element \( q_t \) represents the natural language sub-task instruction provided to the agent, and its embedding \( \mathbf{e}_t = \mathrm{Embed}(q_t) \) is used for semantic similarity retrieval. \( T_t \subseteq \mathcal{T} \) is the subset of tools invoked in the action \( a_t \), with \( T_t = \emptyset \) if no tools are used. The terms \( c_t \) and \( d_t \) denote the monetary cost and wall-clock duration incurred during this step, respectively. \( \sigma_t \in \{0, 1\} \) is a binary indicator of execution success (\textit{e.g.}, whether a tool call executed without error), and \( \mathbb{P}(\tau) \) is the task-level success rate, linking the contribution of step \( t \) to the global task outcome.

After each full task execution, the knowledge base \( \mathcal{K} \) is dynamically updated by appending all newly generated records:
\begin{equation}
\mathcal{K} \leftarrow \mathcal{K} \cup \{ \mathcal{R}_t \}_{t=0}^{T-1}.
\end{equation}
This granular, step-wise logging ensures that \( \mathcal{K} \) evolves into a rich repository capturing the nuanced interplay between sub-task characteristics, model choices, and their resultant impact on performance, cost, and efficiency. This accumulated empirical knowledge forms the foundation for the subsequent retrieval and selection stages.

\subsection{Multi-Faceted Retrieval}

With the knowledge base \( \mathcal{K} \) established, the first active stage of \ourmethod is to retrieve a diverse set of historically analogous records when a new sub-task arises. Given a new sub-task at step \( t' \), characterized by the active agent \( i_{t'} = \mu(t') \) and its instruction \( q_{t'} \), our goal is to gather a comprehensive candidate set \( \mathcal{K}_{\text{cand}} \subseteq \mathcal{K} \). Instead of enforcing a strict intersection of criteria, our approach aggregates records that match on \textit{at least one} of three key facets: agent role, semantic similarity, or tool-use profile. This disjunctive strategy ensures a broad and varied pool of evidence for the subsequent decision-making process.
Formally, the final candidate set \( \mathcal{K}_{\text{cand}} \) is constructed as the union of three independently retrieved subsets:
\begin{equation}
\label{eq:retrieval_union}
\mathcal{K}_{\text{cand}} = \mathcal{K}_{\text{agent}} \cup \mathcal{K}_{\text{sem}} \cup \mathcal{K}_{\text{tool}},
\end{equation}
where each subset is retrieved based on a distinct relevance facet:

% \begin{enumerate}[leftmargin=3em, itemsep=-0.1em]
\noindent\textbf{\ding{228} Agent Role Matching} identifies all historical steps performed by the same agent role, capturing functionally equivalent precedents.
    \begin{equation}
    \mathcal{K}_{\text{agent}} = \{ \mathcal{R}_t \in \mathcal{K} \mid i_t = i_{t'} \}
    \end{equation}

\noindent\textbf{\ding{228} Semantic Similarity Retrieval} retrieves records of sub-tasks that are semantically close to the current one \( q_{t'} \). Using a pre-trained sentence encoder \( \text{Embed}(\cdot) \) and a similarity metric (\textit{e.g.}, cosine similarity), we select all records whose instruction embeddings \( \mathbf{e}_t \) surpass a threshold \( \theta_{\text{sim}} \):
    \begin{equation}
    \small
    \mathcal{K}_{\text{sem}} = \{ \mathcal{R}_t \in \mathcal{K} \mid \text{sim}(\text{Embed}(q_{t'}), \mathbf{e}_t) \geq \theta_{\text{sim}} \},
    \end{equation}
where $\text{Embed}(\cdot)$ is implemented via MiniLM~\citep{wang2020minilm} and $\text{sim}(\cdot,\cdot)$ adopts cosine similarity.

\noindent\textbf{\ding{228}Tool Congruence Retrieval} gathers records based on operational similarity, targeting sub-tasks that likely require similar tool interactions. This facet is specifically designed for instructions $q_{t'}$ where tool usage is anticipated. We first employ a lightweight prediction function, $\text{PredictTools}(q_{t'})$ (whose implementation is detailed in \Cref{app:tool}), to analyze the instruction and infer a set of probable tools, denoted as $T'_{pred}$. A historical record $\mathcal{R}_t$ is then retrieved if its set of invoked tools, $T_t$, has a non-empty intersection with this predicted set:
    \begin{equation}
    \small
    \mathcal{K}_{\text{tool}} = \{ \mathcal{R}_t \in \mathcal{K} \mid T_t \cap \text{PredictTools}(q_{t'}) \neq \emptyset \}.
    \end{equation}
By taking the union of these three sets, \ourmethod assembles a rich and multi-dimensional collection of precedents, \( \mathcal{K}_{\text{cand}} \). This set forms the empirical foundation for the subsequent filtration stage, where we will derive model-specific statistics and identify the Pareto-optimal frontier.

\definecolor{tablepink}{RGB}{253, 231, 231}
\definecolor{tableyellow}{RGB}{254, 243, 215}
\definecolor{tableblue}{RGB}{222, 235, 246}

\begin{table*}[!tbp]
\centering
\caption{Performance comparison on the GAIA and BrowseComp+ benchmarks against both manual and routing-based baselines. For GAIA, results are reported separately for each level, which are defined according to the difficulty of queries by the original benchmark.
} % 更新了标题以反映内容
\vspace{-0.6em}
\label{tab:performance_comparison_1}
\renewcommand{\arraystretch}{1.4}
\renewcommand{\tabcolsep}{2.5pt} % 稍微增加了列间距
\resizebox{\textwidth}{!}{
\begin{tabular}{l l l | ccc ccc ccc ccc | ccc}
\Xhline{1.2pt}
% --- 表格标题行 ---
& \multirow{2}{*}{Setting} & \multirow{2}{*}{LLM} & \multicolumn{3}{c}{\textbf{GAIA (All levels)}} & \multicolumn{3}{c}{\textbf{Level 1}} & \multicolumn{3}{c}{\textbf{Level 2}} & \multicolumn{3}{c}{\textbf{Level 3}} & \multicolumn{3}{c}{\textbf{BrowseComp+}} \\
\cmidrule(lr){4-6} \cmidrule(lr){7-9} \cmidrule(lr){10-12} \cmidrule(lr){13-15} \cmidrule(lr){16-18}
& & & Perf.(\%) & Cost(\$) & Delay(h) & Perf.(\%) & Cost(\$) & Delay(h) & Perf.(\%) & Cost(\$) & Delay(h) & Perf.(\%) & Cost(\$) & Delay(h) & Perf.(\%) & Cost(\$) & Delay(h) \\
\Xhline{1pt}

\multirow{8}{*}{\rotatebox{90}{CK-Pro}}
& \cellcolor{tablepink} 
& \cellcolor{tablepink}Qwen3-14b & 10.91 & 4.16 & 24.67 & 16.77 & 1.06 & 6.23 & 10.34 & 2.58 & 14.90 & 0.00 & 0.52 & 3.54 & 6.50 & 12.50 & 6.80 \\ 
&\cellcolor{tablepink} & \cellcolor{tablepink}Claude-4 & 58.28 & 359.32 & 45.14 & 67.92 & 96.46 & 6.83 & 58.14 & 192.64 & 25.35 & 37.50 & 70.22 & 12.96 & 33.50 & 220.50 & 20.15 \\ 
&\cellcolor{tablepink} & \cellcolor{tablepink}GPT-4o & 42.42 & 82.86 & 35.43 & 59.91 & 27.56 & 7.35 & 39.22 & 43.86 & 21.50 & 15.26 & 11.44 & 6.58 & 24.28 & 113.46 & 9.43 \\ % 
& \cellcolor{tablepink}\multirow{-4}{*}{Manual} & \cellcolor{tablepink}GPT-4.1 & 48.48 & 61.84 & 23.04 & 56.29 & 16.96 & 7.46 & 49.73 & 35.26 & 10.87 & 26.77 & 9.62 & 4.71 & 31.44 & 185.01 & 18.69 \\ 
\cmidrule(l){2-18}
& \cellcolor{tableyellow} 
& \cellcolor{tableyellow}PromptLLM & 47.88 & 128.15 & 57.70 & 58.91 & 31.27 & 10.88 & 47.65 & 80.24 & 30.93 & 24.39 & 16.64 & 15.89 & 26.87 & 133.22 & 17.27 \\ 
& \cellcolor{tableyellow} & \cellcolor{tableyellow}GraphRouter & 46.67 & 116.10 & 53.32 & 58.13 & 28.50 & 10.20 & 46.23 & 72.10 & 28.90 & 22.94 & 15.50 & 14.22 & 27.90 & 125.80 & 16.90 \\ 
& \cellcolor{tableyellow}\multirow{-3}{*}{Routing} & \cellcolor{tableyellow}MasRouter & 53.50 & 120.15 & 52.95 & 67.92 & 27.80 & 9.80 & 49.74 & 77.40 & 28.15 & 23.51 & 14.95 & 15.00 & 28.50 & 128.90 & 17.10 
\\
\cmidrule(l){2-18}
& \cellcolor{tableblue}Ours & \cellcolor{tableblue}\ourmethod & 63.19 & 85.40 & 38.75 & 83.02 & 26.50 & 6.10 & 59.30 & 33.60 & 18.30 & 33.33 & 25.30 & 14.35 & 38.72 & 79.30 & 10.33 \\
\midrule

% --- Smolagent Group ---
\multirow{7}{*}{\rotatebox{90}{Smolagent}}
& \cellcolor{tablepink}
& \cellcolor{tablepink}Qwen3-14b & 11.52 & 0.82 & 8.60 & 15.35 & 0.30 & 3.00 & 4.73 & 0.42 & 1.20 & 27.39 & 0.10 & 4.40 & 5.80 & 1.95 & 4.25 \\
&  \cellcolor{tablepink} & \cellcolor{tablepink}Claude-4 & 46.06 & 76.90 & 29.10 & 56.26 & 21.20 & 12.80 & 46.23 & 41.60 & 14.50 & 22.94 & 14.10 & 1.80 & 28.20 & 85.60 & 16.50 \\ 
& \cellcolor{tablepink} \multirow{-3}{*}{Manual}  & \cellcolor{tablepink}GPT-4.1 & 39.39 & 18.10 & 7.90 & 55.00 & 5.70 & 3.20 & 36.30 & 8.90 & 2.60 & 16.01 & 3.50 & 2.10 & 25.50 & 58.20 & 11.20 \\
\cmidrule(l){2-18}
& \cellcolor{tableyellow}
& \cellcolor{tableyellow}PromptLLM & 40.00 & 17.50 & 12.20 & 49.80 & 4.10 & 6.50 & 40.92 & 8.90 & 3.80 & 15.04 & 4.50 & 1.90 & 23.10 & 45.80 & 10.80 \\ 
& \cellcolor{tableyellow} & \cellcolor{tableyellow}GraphRouter & 38.79 & 15.70 & 11.30 & 47.95 & 3.70 & 6.00 & 39.78 & 8.00 & 3.50 & 15.03 & 4.00 & 1.80 & 24.20 & 42.50 & 10.50 \\ 
& \cellcolor{tableyellow}\multirow{-3}{*}{Routing} & \cellcolor{tableyellow}MasRouter & 40.61 & 17.00 & 12.30 & 49.82 & 4.00 & 6.80 & 42.07 & 8.50 & 3.70 & 15.04 & 4.50 & 1.80 & 24.80 & 44.10 & 10.90 \\ 
\cmidrule(l){2-18}
& \cellcolor{tableblue}Ours & \cellcolor{tableblue}\ourmethod & 56.36 & 15.60 & 8.90 & 73.91 & 3.70 & 2.90 & 53.73 & 6.20 & 4.80 & 27.04 & 5.70 & 1.20 & 32.50 & 25.50 & 9.80 \\ 

\Xhline{1.2pt}
\end{tabular}
}
\end{table*}

\subsection{Pareto-Optimal Filtration and Selection}

Having retrieved a candidate set of historical records \( \mathcal{K}_{\text{cand}} \), the final stage of \ourmethod distills this empirical evidence into a decisive action. The process begins by identifying the unique LLMs, \( \mathcal{L}_{\text{cand}} = \{ l_t \mid \mathcal{R}_t \in \mathcal{K}_{\text{cand}} \} \), that have appeared in the retrieved records. For each candidate LLM \( l \in \mathcal{L}_{\text{cand}} \), we compute its aggregated trilemma metrics by averaging over the relevant subset of records \( \mathcal{K}_{\text{cand}}(l) = \{ \mathcal{R}_t \in \mathcal{K}_{\text{cand}} \mid l_t = l \} \). This yields a statistical profile \( (\hat{\mathbb{P}}(l), \hat{\mathbb{C}}(l), \hat{\mathbb{D}}(l)) \), where $
    \hat{\mathbb{P}}(l) =  \!\!\! \sum_{\tiny \mathcal{R}_t \in \mathcal{K}_{\text{cand}}(l)} {\mathbb{P}(\tau) \cdot \sigma_t}/{|\mathcal{K}_{\text{cand}}(l)|}$,
    $\hat{\mathbb{C}}(l) =   \!\!\!\sum_{\tiny \mathcal{R}_t \in \mathcal{K}_{\text{cand}}(l)} {c_t}/{|\mathcal{K}_{\text{cand}}(l)|}$, and
    $\hat{\mathbb{D}}(l) = \!\!\! \sum_{\tiny \mathcal{R}_t \in \mathcal{K}_{\text{cand}}(l)} {d_t}/{|\mathcal{K}_{\text{cand}}(l)|}$. 
 Subsequently, we perform Pareto filtration. An LLM \( l \) is considered dominated if another LLM \( l' \) exists that is superior or equal on all three axes (higher performance, lower cost, lower duration) and strictly superior on at least one. By retaining only the non-dominated models, we form the Pareto-optimal set, \( \mathcal{L}_{\text{pareto}} \).

A purely greedy selection from \( \mathcal{L}_{\text{pareto}} \) would stifle exploration. To address this, we employ \textit{Thompson sampling}~\citep{russo2020tutorialthompsonsampling}. We treat each of the three trilemma metrics as a continuous variable drawn from a Normal distribution and model the uncertainty over its mean and variance using a Normal-Inverse-Gamma (NIG) conjugate prior.

For each model $l \in \mathcal{L}_{\text{pareto}}$, we dynamically construct its posterior distributions based on the retrieved evidence in $\mathcal{K}_{\text{cand}}(l)$. We first compute the sample statistics for each metric $m \in \{\mathbb{P}, \mathbb{C}, \mathbb{D}\}$: the count $n_l$, the sample mean $\bar{x}_{m,l}$, and the sample variance $s^2_{m,l}$. These statistics are used to parameterize the NIG posteriors, $\text{NIG}(\mu_{m,l}, \nu_{m,l}, \alpha_{m,l}, \beta_{m,l})$, where $\mu_{m,l} = \bar{x}_{m,l}$, $\nu_{m,l}=n_l$, $\alpha_{m,l}=n_l/2$, and $\beta_{m,l}=(n_l-1)s^2_{m,l}/2$, assuming uninformative priors. At decision time, we draw one sample from each model's full posterior for each metric to generate a stochastic realization of its utility. The model \( l^* \) with the highest sampled utility is selected:
\begin{equation}
\small
\label{eq:selection}
\begin{gathered}
    \text{For each } l \in \mathcal{L}_{\text{pareto}} \text{ and each metric } m \in \{\mathbb{P}, \mathbb{C}, \mathbb{D}\}: \\
    (\tilde{\mu}_{m,l}, \tilde{\sigma}^2_{m,l}) \sim \text{NIG}(\mu_{m,l}, \nu_{m,l}, \alpha_{m,l}, \beta_{m,l}) \\
    \tilde{x}_{m,l} \sim \mathcal{N}(\tilde{\mu}_{m,l}, \tilde{\sigma}^2_{m,l}), \\
     \quad U'(l) = w_p \cdot \tilde{x}_{\mathbb{P},l} - w_c \cdot \tilde{x}_{\mathbb{C},l} - w_d \cdot \tilde{x}_{\mathbb{D},l} \\
    l^* = \underset{l \in \mathcal{L}_{\text{pareto}}}{\text{argmax}} \left( U'(l) \right),
\end{gathered}
\end{equation}
where the weights \((w_p, w_c, w_d)\) reflect the desired trilemma trade-off.

Crucially, this selection is not the end of the process. Once the agent powered by \( l^* \) completes its action, the observed outcome is logged back into the knowledge base \( \mathcal{K} \). This closes the feedback loop, ensuring that every decision and its outcome contribute to the system's ever-improving wisdom, thereby realizing the self-evolving nature of \ourmethod.

\section{Experiment}
\vspace{-0.5em}
\subsection{Experiment Setup}
\vspace{-0.4em}
\paragraph{Frameworks.} We select three representative agent frameworks, ordered by increasing architectural complexity: \ding{182} \textbf{ReAct}~\citep{yao2023react}, a single-agent architecture that iteratively alternates between ``observation–action–reasoning''; \ding{183} \textbf{Smolagents}\footnote{\url{https://github.com/huggingface/smolagents}}, a dual-agent framework comprising a manager agent and a tool agent; and \ding{184} \textbf{Cognitive Kernel-Pro}~\citep{fang2025cognitivekernelproframeworkdeep} by Tencent, which achieved open-source SOTA on GAIA in August 2025 and features a hierarchical multi-agent structure (\textit{e.g.}, main agent, web agent, and file agent.)

\begin{table*}[!tbp]
\centering
\caption{Performance comparison on HotpotQA, DS-1000 and DDXPlus benchmarks.}
\vspace{-0.7em}
\label{tab:performance_comparison}
\scriptsize
\renewcommand{\arraystretch}{1.4}
\renewcommand{\tabcolsep}{5.5pt}
\begin{tabular}{ l l | ccc ccc ccc ccc }
\Xhline{1.2pt}

 \multirow{2}{*}{\textbf{Setting}} & \multirow{2}{*}{\textbf{Method/Model}} & \multicolumn{3}{c}{\textbf{DS-1000}} & \multicolumn{3}{c}{\textbf{HotpotQA}} & \multicolumn{3}{c}{\textbf{DDXPlus}} & \multicolumn{3}{c}{\textbf{Avg.}}   \\
\cmidrule(lr){3-5} \cmidrule(lr){6-8} \cmidrule(lr){9-11} \cmidrule(lr){12-14}
 & & Perf. & Cost\$ & Delay(h) & Perf. &  Cost\$ & Delay(h) & Perf. & Cost\$ & Delay(h) & Perf. & Cost\$ & Delay(h)\\
\Xhline{1pt}

 \cellcolor{tablepink} & \cellcolor{tablepink}GPT-4o             & 41.40 & 54.00 & 12.32 & 83.84 & 391.98 & 67.29 & 57.58 & 135.10 & 34.31 & 60.94 & 581.08 & 113.92 \\
 \cellcolor{tablepink} & \cellcolor{tablepink}GPT-4.1            & 29.40 & 49.59 & 12.01 & 87.00 & 213.22 & 78.77 & 55.30 & 89.82 & 21.95 & 57.23 & 352.63 & 112.73 \\
 \cellcolor{tablepink} & \cellcolor{tablepink}Gemini-2.5-pro     & 54.30 & 68.33 & 16.17 & 87.40 & 343.43 & 79.18 & 81.10 & 109.85 & 27.65 & 74.27 & 521.61 & 123.00 \\
 \cellcolor{tablepink} \multirow{-4}{*}{\textbf{Manual}} & \cellcolor{tablepink}Qwen3-14b        & 38.20 & 12.68 & 10.86 & 81.40 & 22.96  & 35.52 & 41.26 & 14.16 & 31.67 & 53.62 & 49.80 & 78.05 \\
 
\cmidrule(l){1-14} 

 \cellcolor{tableyellow} & \cellcolor{tableyellow}PromptLLM        & 52.00 & 26.66 & 11.09 & 85.20 & 47.02  & 62.46 & 60.07 & 127.05 & 26.88 & 65.76 & 200.73 & 100.43 \\
 \cellcolor{tableyellow} & \cellcolor{tableyellow}GraphRouter      & 52.80 & 24.50 & 11.50 & 86.10 & 64.80  & 61.80 & 62.50 & 119.50 & 27.50 & 67.13 & 208.80 & 100.80 \\
 \cellcolor{tableyellow} \multirow{-3}{*}{\textbf{Routing}}      & \cellcolor{tableyellow}MasRouter        & 53.50 & 25.10 & 11.80 & 88.50 & 59.50  & 53.10 & 73.10 & 92.38 & 32.30 & 71.70 & 176.98 & 97.20 \\

\cmidrule(l){1-14}

 \cellcolor{tableblue} \textbf{Ours} & \cellcolor{tableblue} \textbf{\ourmethod}      & 56.50 & 23.50 & 11.20 & 87.80 & 49.10 & 60.50 & 79.50 & 65.80 & 20.53 & 74.60 & 138.40 & 92.23 \\
 
\Xhline{1.2pt}
\end{tabular}
% }
\vspace{-0.8em}
\end{table*}

\vspace{-0.6em}
\paragraph{Baselines.} We compare against two main categories of baselines: \textbf{(i) manual setting}, which includes each framework’s original predefined LLM configuration as well as manually specified setups (\textit{e.g.}, uniformly using \llmname{gpt-4.1} or \llmname{Gemini-2.5-pro}). The choice of manually assigned LLMs may vary across frameworks depending on compatibility requirements. \textbf{(ii) model routing}, encompassing SOTA routing methods like {PromptLLM}~\citep{feng2024graphroutergraphbasedrouterllm}, {MasRouter}~\citep{yue2025masrouter}, and {GraphRouter}~\citep{feng2024graphroutergraphbasedrouterllm}\footnote{Our code will be available at \url{https://github.com/bingreeky/evo-route}.}.

\vspace{-0.6em}
\paragraph{Method Configurations.} We fill our LLM pool $\mathcal{L}$ with models of varing prices and sizes: \{\llmname{Qwen-14b}, \llmname{GPT-4.1}, \llmname{GPT-4o}, \llmname{Claude-4-Sonnet} (\llmname{(Claude-4)}), \llmname{Gemini-2.5-Flash}, \llmname{Gemini-2.5-Pro}\}. The model prices are listed in \Cref{app:price}.

\vspace{-0.5em}
\paragraph{Parameter Configurations.}
For our multi-faceted retrieval, we set the semantic similarity threshold $\theta_{\text{sim}}$ to 0.85 to balance retrieval recall and precision. In the selection stage, we leverage uninformative NIG priors ($\nu_0=0, \alpha_0=0.5, \beta_0=0$) for each metric. The final selection is guided by a utility function with manually configured weights $(w_p, w_c, w_d) = (1.0, 0.1, 0.05)$, which prioritizes performance while still penalizing cost and delay.

\vspace{-0.3em}
\paragraph{Benchmark and Evaluations.} We evaluate our approach on five benchmarks. Two are deep research benchmarks: \textbf{GAIA}~\citep{mialon2023gaia}, a widely used general-purpose agentic benchmark that assesses capabilities such as file reading, web browsing, and coding, and is divided into three tiers based on difficulty; and \textbf{BrowseComp+}~\citep{chen2025browsecompplusfairtransparentevaluation}, an enhanced version of OpenAI’s BrowseComp~\citep{wei2025browsecompsimplechallengingbenchmark} that corrects erroneous cases and provides more stable evaluations. The remaining three are from StreamBench~\citep{wu2024streambenchbenchmarkingcontinuousimprovement}: \textbf{DS-1000}~\citep{lai2022ds1000naturalreliablebenchmark} for data science tasks, \textbf{HotpotQA}~\citep{yang2018hotpotqa} for web search, and \textbf{DDXPlus}~\citep{tchango2022ddxplusnewdatasetautomatic} for medical reasoning. The statistics of above datasets are placed in \Cref{app:dataset}.

\subsection{Main Results}\label{sec:main-result}
\begin{figure}[!t]
\vspace{-0.3em}
  \begin{center}
    \includegraphics[width=\columnwidth]{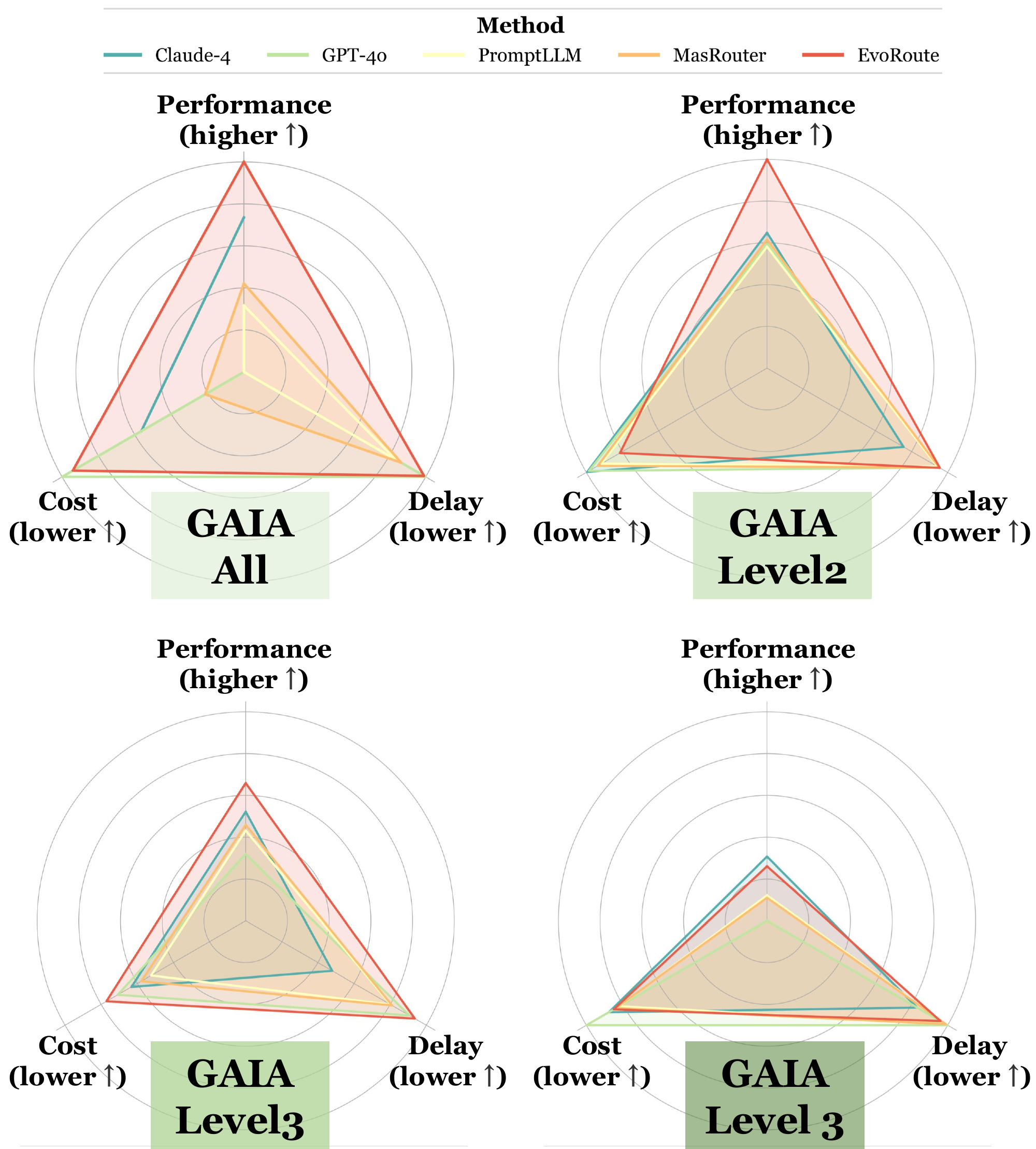}
  \end{center}
  \vspace{-0.2cm}
  \caption{Comparative analysis across three key metrics: performance, cost, and delay, on all subsets of GAIA. All metrics are globally normalized, and values for cost/delay are inverted, such that a larger enclosed area signifies better economy/efficiency.}
  \label{fig:tri}
  \vspace{-0.4cm}
\end{figure}

\vspace{-0.3em}
\Cref{tab:performance_comparison,tab:performance_comparison_1} present the comparison of \ourmethod against different baselines across three key dimensions: \textit{performance}, \textit{cost}, and \textit{latency}. \Cref{fig:tri} further illustrates a radar visualization of results across different levels of GAIA. Our main findings are summarized as follows.

\paragraph{Existing Agent Systems Grapple with the Performance-Efficiency-Economy Trilemma.}
Our empirical analysis reveals that contemporary agentic systems fundamentally struggle to reconcile performance, cost, and latency. Achieving state-of-the-art performance typically incurs exorbitant costs and significant delays. For instance, using the powerful \llmname{Gemini-2.5-Pro}+ReAct achieves a leading $74.27\%$ average performance, but at a prohibitive cost of $\$521.61$ and a delay of $123$ hours, as shown in \Cref{tab:performance_comparison}. Similarly, \llmname{Claude-4}+CK-Pro reaches a strong $55.15\%$ on GAIA, yet demands an unsustainable $\$359.32$ (\Cref{tab:performance_comparison_1}). Conversely, opting for economical models like \llmname{Qwen3-14b}+CK-Pro drastically reduces cost to just $\$4.16$, but at the expense of a sharp performance drop to $10.91\%$. While existing SOTA routing methods like MasRouter offer some mitigation (achieving $71.70\%$ average performance for $\$176.98$), they provide only incremental improvements. These methods still lag behind top-performing models in accuracy and fail to fundamentally break the trade-off.

% \begin{wrapfigure}{r}{0.4\textwidth}

\paragraph{\ourmethod Effectively Alleviates the Trilemma.} 
In contrast, \ourmethod demonstrates a remarkable ability to navigate the trilemma, consistently achieving state-of-the-art performance while substantially reducing cost and latency. Integrated with CK-Pro, it surpasses the strong \llmname{Claude-4} baseline on GAIA ($63.18\%$ vs. $58.28\%$), cuts operational costs by over $76\%$ ($\$85.40$ vs. $\$359.32$), and maintains comparable latency. On BrowseComp+, it similarly outperforms \llmname{Claude-4} ($38.72\%$ vs. $33.50\%$) at under $36\%$ of the cost ($\$79.30$ vs. $\$220.50$) and roughly half the latency ($10.33$h vs. $20.15$h). Across all tested configurations (\Cref{tab:performance_comparison}), \ourmethod achieves the highest average performance ($74.60\%$), slightly exceeding \llmname{Gemini-2.5-Pro} ($74.27\%$), while incurring only $26\%$ of the cost ($\$138.40$ vs. $\$521.61$) and reducing latency by $25\%$ ($92.23$h vs. $123.00$h). These results firmly establish \ourmethod as an effective solution to the agent system trilemma.

\begin{figure}[!t]
\vspace{-0.3em}
  \begin{center}
    \includegraphics[width=\columnwidth]{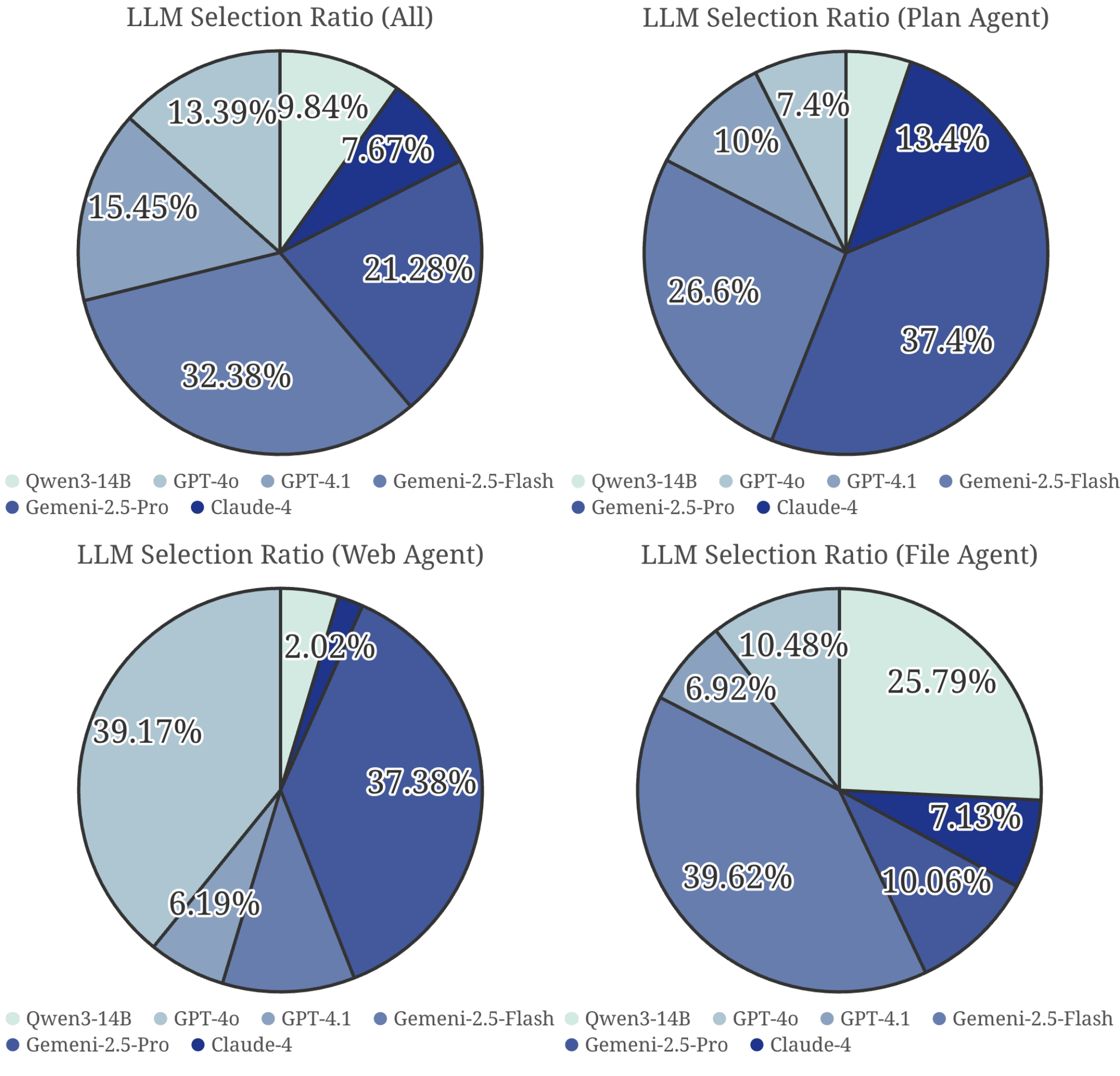}
  \end{center}
  \vspace{-0.3cm}
  \caption{LLM selection distribution of \ourmethod+CK-Pro across different agent roles.}
  \label{fig:pie}
  \vspace{-0.5cm}
\end{figure}

\vspace{-0.3em}
\subsection{Framework Analysis}

\vspace{-0.3em}
\paragraph{Visualization.}
We analyze \ourmethod's LLM selection across agent roles on CK-Pro+GAIA (\Cref{fig:pie}). The results indicate that model allocation aligns with sub-task complexity: cognitively demanding tasks handled by the \texttt{Plan Agent} favor high-capability models (\texttt{Gemini-2.5-Pro} $37.4\%$, \texttt{Claude-4} $13.4\%$), whereas simpler, operational tasks of the \texttt{File Agent} predominantly use cost-efficient models (\texttt{Qwen3-14B} $25.79\%$), with premium models like \texttt{Gemini-2.5-Pro} and \texttt{Claude-4} dropping to $10.06\%$ and $7.13\%$. This shows that \ourmethod strategically reserves its most powerful (and expensive) models for critical reasoning, while employing efficient models for routine sub-tasks.

\begin{table}[h]
\centering
\caption{Ablation study of four variants, each disabling a key component (tested on GAIA Level 1). }
\vspace{-0.6em}
\label{tab:ablation}
\renewcommand{\arraystretch}{1.3}
\resizebox{\linewidth}{!}{%
\begin{tabular}{l|ccc}
\Xhline{1.2pt}
\textbf{Method Variant} & \textbf{Perf. (\%)} & \textbf{Cost (\$)} & \textbf{Delay (h)} \\
\Xhline{1pt}
\rowcolor{tableblue!25} 
\ourmethod (Full Model) & \textbf{83.02} & \textbf{26.50} & \textbf{6.10} \\
\midrule
w/o $\mathcal{K}$ cold start & 69.81 \tiny{(-13.21)} & 29.50 & 6.30 \\
w/o Multi-Faceted Retrieval & 71.70 \tiny{(-11.22)} & 32.10 & 6.95 \\
w/o Thompson Sampling & 76.50 \tiny{(-6.52)} & 29.80 & 6.40 \\
w/o Pareto Filtration & 81.50 \tiny{(-1.52)} & 28.20 & 6.35 \\
\Xhline{1.2pt}
\end{tabular}%
}
\vspace{-1.5em}
\end{table}

\vspace{-0.5em}
\paragraph{Ablation Study}
We performed an ablation study on GAIA Level~$1$ to validate our core components (\Cref{tab:ablation}). We evaluate four variants: \textbf{w/o $\mathcal{K}$ cold start} (starting with an empty knowledge base), \textbf{w/o Multi-Faceted Retrieval} (using only semantic similarity), \textbf{w/o Thompson Sampling} (using a greedy policy), and \textbf{w/o Pareto Filtration} (omitting model pruning). The absence of a bootstrapped knowledge base is the most detrimental, causing performance to plummet by $13.21\%\downarrow$. Disabling the multi-faceted retrieval is nearly as damaging, resulting in an $11.22\%$ performance decrease and incurring the highest cost (\$$32.10$) and delay ($6.95$\,h) among all variants. Furthermore, replacing Thompson sampling with a greedy policy reduces performance by $6.52\%$, while omitting Pareto filtration yields a smaller $1.52\%$ drop, confirming that both principled exploration and efficiency-focused pruning are vital to the system's overall effectiveness.

\vspace{-0.5em}
\paragraph{Architectural Overhead on Simpler Tasks.} 
One may note that, in \Cref{sec:main-result}, ReAct is evaluated on simpler tasks (DS-1000, HotpotQA), while Smolagent and CK-Pro are reserved for GAIA and BrowseComp+. This reflects a practical observation: deploying more complex frameworks on simple tasks can be \textit{counterproductive}. For example, we initially discover that \llmname{Gemini-2.5-Pro}+Smolagent scored only $31.7\%$ on DS-1000, below the $38.2\%$ of the simpler \llmname{Qwen3-14b}+ReAct. Trace analysis suggests that Smolagent’s features (\textit{e.g.}, web search and high-level planning) introduce overhead by overcomplicating tasks solvable with a single LLM I/O. We term this ``architectural overfitting,'' where structural complexity and specialized tools hinder rather than help. This highlights that, the optimal agent architecture should match the task’s intrinsic complexity, not its maximal sophistication.

\vspace{-0.5em}
\section{Conclusion}
\vspace{-0.5em}
In this work, we introduced \ourmethod, a dynamic model routing framework designed to systematically address the agent system trilemma.  Our experiments on challenging benchmarks like GAIA and BrowseComp+ empirically validate this paradigm, delivering substantial reductions in monetary cost (up to $80\%$) and latency (over $70\%$) without compromising task success. Ultimately, \ourmethod represents a crucial step towards making powerful agentic AI systems more practical, scalable, and economically viable for real-world deployment.

\section*{Limitation \& Ethical Concerns}  
Our evaluation focuses on CK-Pro and Smolagent and does not select other deep-research frameworks such as OWL, Agent-Orchestra, or AIME. Given the proliferation of deep research systems and their broadly similar architectural paradigms—typically consisting of a central coordinator supported by multiple specialized sub-agents—we believe our experiments already capture a representative spectrum of agentic systems from simple to complex. Moreover, the substantial token costs associated with these systems (as shown in \Cref{tab:performance_comparison_1}) make exhaustive large-scale benchmarking impractical. Regarding ethical considerations, since our study relies exclusively on standard public benchmarks and widely used open frameworks, we identify no immediate ethical risks associated with this work.

\section*{Contributions}
G. Zhang was primarily responsible for the method implementation, manuscript preparation, visualization, and experimental analysis. K. Yang conducted a substantial portion of the experimental work and was also one of the core contributors. H. Yu and B. Wu provided extensive guidance and engaged in in-depth discussions throughout the project. F. Huang, Y. Li, and S. Yan offered senior-level supervision and strategic guidance.

\bibliography{custom}

\appendix

\section{Cold Start Issue}\label{app:cold}

A fundamental challenge for any experience-based system is the ``cold start'' issue: the experience base, $\mathcal{K}$, is initially empty (\textit{tabula rasa}), rendering the retrieval and routing mechanisms ineffective as they have no prior data upon which to base their decisions. To address this, we implement a dedicated exploration strategy designed to populate $\mathcal{K}$ with a diverse and informative set of initial experiences before the system is deployed for operational use.

We leverage a curated set of 50 agentic tasks from the TaskCraft dataset~\citep{shi2025taskcraftautomatedgenerationagentic}. For each of these 50 tasks, we execute both the Smolagent and CK-Pro frameworks from start to finish. We employ a stochastic exploration policy to maximize the diversity of the collected data. Specifically, at each step $t$ of a task's execution, the LLM backbone $l_t$ for the active agent $i_t = \mu(t)$ is selected via uniform random sampling from the entire pool of available models $\mathcal{L}$. Formally, the selection is made as follows:
\begin{equation}
l_t \sim \mathcal{U}(\mathcal{L}),
\end{equation}
where $\mathcal{U}(\mathcal{L})$ denotes the uniform distribution over the discrete set of LLM backbones.

Upon the completion of each of the 50 tasks, the full execution trajectory $\tau$ is processed as described in our main methodology. The complete set of step-wise records $\{\mathcal{R}_t\}_{t=0}^{T-1}$ is extracted and used to populate the initially empty experience base $\mathcal{K}$. 

We further emphasize that the cold-start process is not costly: it required only $\$28.8$ and approximately $3.6$ hours to complete, yielding around $480$ step-level records that provide a sufficiently informative prior. Overall, this initialization overhead remains modest.

\section{Tool Prediction Function}\label{app:tool}

The tool prediction function \texttt{PredictTools($q_t$)} employs a two-stage hybrid strategy to balance predictive accuracy with minimal computational overhead. Initially, the function performs a near-instantaneous heuristic check, using a predefined dictionary to map explicit trigger keywords (e.g., ``search'' for \texttt{web\_search}; ``run'', ``plot'' for \texttt{code\_interpreter}) to their corresponding tools. This handles the majority of clear-cut cases with zero latency or API cost. If, and only if, this initial heuristic fails to find a match, the function escalates to a more powerful fallback: a single API call to a cheap and effective LLM (practically, \llmname{Qwen3-14b}). This model is prompted in a zero-shot manner to analyze the instruction's semantics and identify the necessary tools from the available set.

\section{Model Price}
\label{app:price}

\begin{table}[!h]
\centering
\small
\caption{The pool of LLM backbones ($\mathcal{L}$) used in our experiments, along with their respective pricing per one million tokens. The models were selected to cover a diverse range of capabilities and operational costs.}
\label{tab:model_prices}
\renewcommand{\arraystretch}{1.2}
\begin{tabular}{lrr}
\toprule
\textbf{Model} & \makecell{Input Price\\(\$/M)} & \makecell{Output Price\\(\$/M)} \\
\midrule
\llmname{Gemini-2.5-Flash}          & \$0.3  & \$2.5  \\
\llmname{Qwen3-14B}          & \$0.05  & \$0.22  \\
\llmname{Claude-4}          & \$3.00  & \$15.00 \\
\llmname{GPT-4o}                    & \$2.50  & \$10.00 \\
\llmname{Gemini-2.5-Pro}           & \$1.25  & \$10.00 \\
\llmname{GPT-4.1}             & \$2.00 & \$8.00 \\
\bottomrule\\
\end{tabular}
\end{table}

 \section{Dataset Details}
\label{app:dataset}
The GAIA benchmark~\citep{mialon2023gaia} offers a broad evaluation suite for general-purpose AI assistants, comprising 165 tasks systematically organized into three difficulty tiers: 53 basic tasks (Level~1), 86 intermediate tasks (Level~2), and 26 advanced tasks (Level~3). The BrowseComp+ benchmark includes 830 evaluation instances, while DS-1000 consists of 1,000 tasks. HotpotQA contains 7,405 queries. For DDXPlus, originally comprising approximately 130K samples, we randomly subsample 1,000 instances for evaluation.

\section{Use of AI Assistants}  
We employed AI-based tools, including large language models, to support various stages of manuscript preparation. These tools were used for language polishing, improving clarity and readability, formatting references, and generating visualizations.

\end{document}